\documentclass{article}

\usepackage[english]{babel}

\usepackage[letterpaper,top=2cm,bottom=2cm,left=3cm,right=3cm,marginparwidth=1.75cm]{geometry}

\usepackage{amsmath}
\usepackage{amssymb}
\usepackage{amsthm}
\usepackage{graphicx}
\usepackage[colorlinks=true, allcolors=blue]{hyperref}
\usepackage{algorithmic}

\usepackage[ruled,vlined,noline,linesnumbered]{algorithm2e}

\usepackage{titlesec}
\usepackage[titletoc]{appendix}
\usepackage{varwidth}

\title{Empirical Bayes for Dynamic Bayesian Networks Using Generalized Variational Inference}
\author{Vyacheslav Kungurtsev, Apaar$^*$, Aarya Khandelwal$^*$, Parth Sandeep Rastogi$^*$\footnote{These authors contributed equally},\\ Bapi Chatterjee, Jakub Mare{\v c}ek }
\date{April 2024}

\begin{document}

\maketitle

\section{Introduction}
Dynamic Bayesian Networks (DBNs) are a class of Probabilistic Graphical Models that enable the modeling of a Markovian dynamic process through defining the kernel transition by the DAG structure of the graph found to fit a dataset. There are a number of structure learners than enable one to find the structure of a DBN to fit data, each of which with its own set of particular advantages and disadvantages. 

The structure of a DBN itself presents transparent criteria in order to identify causal discovery between variables. However, without the presence of large quantities of data, identifying a ground truth causal structure becomes unrealistic in practice. However, one can consider a procedure by which a set of graphs identifying structure are computed as approximate noisy solutions, and subsequently amortized in a broader statistical procedure fitting a mixture of DBNs. Each component of the mixture presents an alternative hypothesis on the causal structure. From the mixture weights, one can also compute the Bayes Factors comparing the preponderance of evidence between different models. 

This presents a natural opportunity for the development of \emph{Empirical Bayesian} methods. See, e.g.~\cite{robbins1992empirical} for a classic overview. Empirical Bayes attempts to combine the frequentist and Bayesian modeling approaches by using a frequentist point estimate as a prior for a subsequent Bayesian model. In this case, we perform this as a mixed integer programming problem~\cite{bartlett2017integer}, solving for the optimal structure and weights on a subsample of the data. This provides us with a point estimate of an optimal (and consistent, see~\cite{kucukyavuz2023consistent}) structure. Subsequently we repeat the procedure with a different subsample of the data while encouraging model heterogeneity. After obtaining $M$ models with least-squares optimized weights $\{\Theta_m\}_{m=1,...,M}$, now we solve a measure valued problem wherein we relax the structure component into learning a mixture across these networks, corresponding to optimization over an $M$ dimensional simplex and at the same time, solving for the measure defining the posterior distribution of the weights using a prior that centers on the MIP solution and enforces an f-divergence from this prior as a constraint.

There are a number of fundamental advantages to this procedure. As described in~\cite{friedman2000being}, in the common scenario of interest for learning (D)BNs, with high dimensionality of the feature space and a paucity of data samples, overconfidence in regards to certainty of structure is often deleterious to appropriate certification of model accuracy and variance. On the other hand, appropriately constructing a mixture model a priori is methodologically challenging as far as the poor scaling of any appropriate encoding with the problem dimension. With an uninformative prior, mixing times can be unreasonably long. However, by initially finding a set of distinct high quality models to serve as the basis of the mixture, we manage to perform Bayesian learning, that is uncertainty quantification for non-asymptotic sample size circumstances, without the potential computational expense of encoding and sampling the entire DBN representation at once. 


This formulation leads to a class of problems of contemporary interest in Bayesian Statistics described as Generalized Variational Inference (GVI), e.g.~\cite{knoblauch2019generalized}. GVI presents a novel probabilistic approach to considering Bayesian statistical models that generalizes classical Variational Inference to consider the use of alternative score functions as well as a generic set of probability distances and divergences. GVI was found to outperform classical Bayesian methods for circumstances of incorrect or imprecise model and prior specification. 

As this statistical method is new while at the same time quite general, few algorithms and procedures have been developed for it. Recently the work~\cite{javeed2023risk} appeared which presented a comprehensive method for a very similar Coherent GVI (CGVI) problem, with theoretical convergence guarantees and promising illustrative performance across a variety of problems and choice of divergences. The procedure transforms the measure valued optimization into a low dimensional convex (however, non-Lipschitz) problem and high dimensional sampling.

In order to simplify the computational task of sampling from high dimensional but low sample settings, we make the simplifying assumption of a linear Structural Equation Model (LSEM). Formally, we have a state vector $X_t$ which evolves as,
\begin{equation}\label{eq:dbnmodel}
    X_t=X_t W+X_{t-1}A_1+...+X_{t-p}A_p+Z
\end{equation}
where $W$ is the adjacency matrix of an acyclic graph for intra-slice edges and $A_i$ for inter-slice edges, respectively, and $Z$ is the noise. Consider the availability of data in the form $\{X_{m,t}\}$ with $m=1,...,M$ independent samples of trajectory length $t=1,...,T$ of dimensionality $X_{m,t}\in\mathbb{R}^d$. This is a popular representation for a DBN, with the sparsity (zero and nonzero) structure of the adjacency matrices corresponding to the presence or lack of edges, thus defining the graph structure of the DBN transitions.

We now proceed to describe the major components of the approach and details of the procedure.

\section{Empirical Bayesian Model and Prior Estimate with MIP}
Learning BNs has a natural association to Integer Programming (IP) and other combinatorial solvers, given the extensive rich combinatorial structure~\cite{cussens2017bayesian}. However, in many cases of interest, the data are high-dimensional and yet we are limited as far as quantities of data samples. In such circumstances, enforcing sparsity becomes necessary to establish meaningful models. At the same time, the statistical regime is not particularly favorable for recovery of an exact sparse solution~\cite{wainwright2019high}. Thus, a discrete programming solution could be too rigid in its ability to fit a wide range of problem instances.

Empirical Bayes presents an approach that attempts to incorporate the advantages of uncertainty quantification of estimates that comes with Bayesian data analysis while mitigating the risks associated with misspecified priors. The procedure is simple: perform standard frequentist likelihood estimation to obtain a point estimate for the model, then subsequently design a prior that is centered on the estimate and includes a ball of uncertainty defined by probability distance. See, e.g.~ \cite{carlin2008bayesian} and \cite{efron2012large} for comprehensive monographs. 

In standard Bayesian approaches, one would present an explicit prior for $(\Theta,\Xi)$ as arising from some external parametrically defined distribution, that is $(\Theta,\Xi)\sim p(\Theta,\Xi;\eta)$ for some parameter $\eta$. A posterior estimate is computed with $p\left((\Theta,\Xi)\vert \{X_{n,t}\},\eta\right)\propto p\left(\{X_{n,t}\}\vert (\Theta,\Xi)p(\Theta,\Xi;\eta\right)$. Typically, however, there is no reason to suspect that $\eta$ is known. In a fully Bayesian approach we would assign a hyperprior distribution $p(\eta)$ and require integration, i.e. treating $\eta$ as a ``nuisance parameter'', by $p\left((\Theta,\Xi)\vert \{X_{n,t}\}\right)\propto \int p\left(\{X_{n,t}\}\vert (\Theta,\Xi)p(\Theta,\Xi;\eta\right)p(\eta)d\eta$. This presents the need for significant computation. Furthermore, with a poorly specified hyperprior $p(\eta)$, especially in the small-sample case we are interested in here, could have an outsized negative influence on the accuracy of the posterior for $(\Theta,\Xi)$. With empirical Bayes, one instead computes an estimate of $\hat{\eta}$ from the data, and uses this to subsequently develop the full model. Note that it is also possible to keep the prior $p((\Theta,\Xi))$ non-parametric, however, in~\cite{carlin2008bayesian} it is noted that parametric estimates of the hyperprior perform better, especially in the far from asymptotic case. For instance one can use the maximum posterior estimate $\hat{\eta}$ given the data, and use $\delta_{\hat{\eta}}$ as the hyperprior distribution for $\eta$, obviating the need for computing a complicated integral. 

We specify a \emph{data driven robust prior} (also referred to as a Parzen window), as inspired from distributionally robust optimization, which has been observed to yield favorable generalization properties for machine learning models~\cite{staib2019distributionally}. To this end we consider that that true Bayesian model corresponds to a mixture of structures together with an uncertainty ball centered on a particular parameter set. 

To this end, we present some important notation:
\begin{equation}\label{eq:datanotation}
\begin{array}{l}
    X_{n,t}\in\mathbb{R}^d,\text{ Data sample of }n'th\text{ trajectory and }t\text{ time instance} \\
    \Theta_m\in \mathbb{R}^{d(\Xi_m)},\,\Xi_m=(\Xi_m^W,
    \Xi_m^A)\in\{0,1\}^{d\times d},\{0,1\}^{d\times d\times p} ,\,m\text{'th model decision variables} \\
    \mathbf{S}(\{X_{t,n}\}_{N,T},S)= \mathcal{S}:=\{X_{i,t}\}_{i=n^1,...,n^S,t\in[T]},\,\{n^1,...,n^S\}\sim \mathcal{U}\left\{\begin{array}{c} N \\ S \end{array}\right\} \\
    (\Theta,\Xi)=\mathop{MIP}(\{X_{n',t}\}),\text{ the MIP solution operation on the data }\{X_{n',t}\}_{n'\in \mathcal{S}}
\end{array}
\end{equation}
Here $\mathbf{S}$ represents the uniform sub-sampling operator. We can now present the generic form of the source of our initial frequentist estimate. 


\subsection{Integer Programming Estimates}

In this case, we perform hierarchical inference based on the mixed nonlinear integer programming formulation defined in~\cite{manzour2021integer} (see also~\cite{kucukyavuz2023consistent}, which presents a relaxation technique using SOCPs for solving large instances).

To this end the DBN model incorporates a directed acyclic graph (DAG) $\bar{\mathcal{G}}=\mathcal{G}(\mathcal{V},\mathcal{E}_W)\cup \mathcal{G}(\mathcal{V},\mathcal{E}_A)$ which defines the present connections in the model. That is $e=\{X_1,X_2\}\in\mathcal{E}_W$ if $W_{1,2}\neq 0$, that is $1$ is a parent of $2$ in $\mathcal{E}_W$.

We present the IP problem for solving $(E_W,E_A)\in \left[\{0,1\}^{d}\times\{0,1\}^{d}\right]\times \left[\{0,1\}^{d}\times\{0,1\}^{d}\right]^p$, below. Note that with integer variables, one can induce a regularization of, effectively $\|W\|_0$, rather than the one norm as previously defined in the one-shot formulation. See the discussion on ~\cite[pg 6]{kucukyavuz2023consistent} indicating that for DAGs, the relaxed $l1$ formulation is often inadequate to enforce sparsity. 
\begin{equation}\label{eq:ip}
    \begin{array}{rl}
\min\limits_{(E_W,E_A,W,A)} & \mathbf{E}(E_W,E_A,W,A;\tilde{\mathcal{S}})+\lambda_W \|E_W\|_0+\lambda_A\|E_A\|_0 \\ &:=\sum\limits_{n=1}^{\tilde{N}} \sum\limits_{t=1}^T\sum\limits_{i=1}^d\left([X_{n,t}]_i-\sum\limits_{j=1}^d W_{j,i}[X_{n,t}]_j\right.\\&\left.-\sum\limits_{l=1}^p\sum\limits_{j=1}^n A_{l,j,i} [X_{n,t-l}]_j\right)^2+\lambda_W \sum\limits_{i,j} [E_W]_{i,j}+\lambda_A\sum\limits_{l,i,j}[E_A]_{l,i,j} \\
\text{s.t. } &  W\cdot (1-E_W)=0,\\
& DAG(E_W), \\
& (E_W,E_{\tilde{W}})\in \left[\{0,1\}^{d^2}\right]\times \left[\{0,1\}^{d^2}\right]\\
& (E_A,E_{\tilde{A}}) \in \left[\{0,1\}^{d^2}\right]^p\times  \left[\{0,1\}^{d^2}\right]^p \\
& W\in\mathbb{R}^{d\times d},\, A\in\mathbb{R}^{p\times d\times d}
    \end{array}
\end{equation}

Thus, we compute the initial set of frequentist point estimates for the structure and parameters $\{\Theta_m,\Xi_m\}$ by:
\begin{algorithm}
\begin{algorithmic}
    \FOR{$m=1,...,M$}
    \STATE Sample $\mathcal{S}^m:=\{X_{n_i^m,\cdot}\} \sim \mathbf{S}\left(\{X_{n,\cdot}\},S\right)$, $n_i^m\in[N]$
    \STATE Solve~\eqref{eq:ip} with $\tilde{\mathcal{S}}
\rightarrow \mathcal{S}^m$ to obtain $(\Theta_m,\Xi_m):=((W^m,A^m),(E_W^m,E_A^m))$, representing the optimal parameters and structure the IP solver found for subsample $m$. 
    \ENDFOR
\end{algorithmic}
\caption{Initial Integer Programming Solutions}\label{alg:initip}
\end{algorithm}

There are other approaches that can also generate a set of, rather than one,  model structure, e.g.~\cite{liao2019finding}. We aimed to explicitly limit the number of structures while encouraging variety in the strategy proposed here. 


\section{Generalized Variational Inference Optimization Problem}
Consider now that we have obtained a set of $M$ candidate Dynamic Bayesian Network structures and their associated parameters $\{\Theta_m\}_{m=1,...,M}$. In order to define our empirical Bayes approach, we define a \emph{Coherent Generalized Variational Inference} (CGVI) problem, which takes the form~\cite{javeed2023risk},
\begin{equation}\label{eq:cgvi}
    \min\limits_{P\in \mathcal{D}}\mathbb{E}_P[\mathbf{E}_N(\Theta)]
\end{equation}
wherein $P$ is a probability measure on a set of paremters $\Theta$, $\mathcal{U}$ is a set of admissible measures. Finally $\mathbf{E}_N$ is a loss function on the $N$ data samples with a model parametrized by $\Theta$, that is
\begin{equation}\label{eq:en}
\mathbf{E}_N(\Theta,\Xi) = \mathbf{E}(\Xi_W,\Xi_A,\Theta_W,\Theta_A;\mathcal{S})
\end{equation}
where we shall sometimes suppress the structure in the case of clear context.

\subsection{Background}
The determination of $\mathcal{D}$ is defined using $\phi$-divergences, a natural measure for probability distances often used for risk measures. Defining a utility functional $\mathcal{U}$ on a space of random variables $\mathcal{M}$, with the disutility functional $\mathcal{V}(X)=-\mathcal{U}(-X)$ a divergence is $\Phi(p_{\Pi})=\mathcal{V}^*(p_{\Pi})$ where $p_{\Pi}$ is the density of the prior measure $\Pi(\Theta)$ and $\mathcal{V}^*$ refers to the Fenchel conjugate of $\mathcal{V}$. 

The application of duality yields the two-dimensional convex problem:
\begin{equation}\label{eq:cgvioptgeneral}
    \min\limits_{\lambda\ge 0,\mu\in\mathbb{R}} \, \left\{\mu+\epsilon\lambda+(\lambda\Phi)^*(\mathbb{E}_N-\mu)\right\}
\end{equation}
with $(\lambda\Phi)^*$ being the Fenchel conjugate of $\lambda\Phi$. 

In our case we will use the R{\' e}nyi divergence, defined as,
\[
D_{\alpha}(p\vert\vert\pi) \triangleq \frac{1}{\alpha-1}\log \mathbb{E}\left[p^{\alpha}\right]
\]
for some $\alpha>0$. Note that when $\alpha=1$ this recovers the KL divergence. Letting $\beta=\frac{\alpha}{\alpha-1}$, this divergence is associated, through the Orlitz conjugate, with the isoelastic disutility function,
\[
v(x)=\left(1+\frac{x}{\beta}\right)^{\beta}-1
\]

With a solution $\bar{\lambda}$, $\bar{\mu}$ to~\eqref{eq:cgvioptgeneral},
the CGVI density associated with the solution is given by:
\begin{equation}\label{eq:posteriorcgvi}
    p(\theta) = \left(1+\frac{\mathbf{E}_N(\theta)-\bar{\mu}}{\beta\bar{\lambda}}\right)^{\beta-1}
\end{equation}
Which we can re-write in Gibbs form, to facilitate sampling with Langevin and Hamilton Monte Carlo,
\begin{equation}\label{eq:posteriorcgvigibbs}
    p(\theta) = \exp\left[\log\left\{\left(1+\frac{\mathbf{E}_N(\theta)-\bar{\mu}}{\beta\bar{\lambda}}\right)^{\beta-1}\right\}\right]=\exp\left\{
    (\beta-1)\log\left[1+\frac{\mathbf{E}_N(\theta)-\bar{\mu}}{\beta\bar{\lambda}}\right]\right\}
\end{equation}

\subsection{CGVI Formulation of Dynamic Bayesian Network Learning}
Now that we have introduced the general CGVI problem, we can proceed to describe how we intend to incorporate it into our general Empirical Bayes procedure.

Recall that we have obtained frequentist estimates for the structure and parameters $\{\Theta_m,\Xi_m\}$ to serve as a pivot for the Empirical Bayes. To this end, we consider that we seek a solution to a hierarchical Bayesian problem. In particular we define a mixture $\mathbf{m}\in \Delta$ that samples from the $M$ structures. Then, we shall see that we can treat the rest as a set of $M$ uncoupled CGVI problems that sample the distribution of weights $\Theta$ for the $m$'th model

Define the loss function over the entire dataset, as a function of a vector of parameters:
\begin{equation}\label{eq:loss}
    \mathbf{E}_N(\phi^m(\theta)):=\mathbf{E}_N(\Xi_m,W(\phi^m_W(\theta)),A(\phi^m_A(\theta))) = \sum\limits_{n=1}^N\sum\limits_{t=1}^T\sum\limits_{i=1}^d\left([X_{n,t}]_i-\sum\limits_{j=1}^d W_{j,i}[X_{n,t}]_j-\sum\limits_{l=1}^p\sum\limits_{j=1}^d A_{l,j,i} [X_{n,t-l}]_j\right)^2
\end{equation}
See that the structure $\Xi_m$ significantly restricts the search space of $(W,A)$. In order to set up the appropriate definitions below for sampling, we introduce the following maps:
\begin{equation}\label{eq:supportWA}
\begin{array}{l}
    \phi^m: \mathbb{R}^{s_m}\rightarrow (\mathbb{R}^{d\times d},\mathbb{R}^{d\times d\times p}),\\
    \mathcal{I}_m = \text{supp}(\Xi_m):=\left\{(i,j)\in [E_A^m]_{i,j}= 1\}\cup\{(i,j):[E_W^m]_{i,j}= 1\}\right\}\\
    s_m = \vert\text{supp}(\Xi_m)\vert,\\
    \phi^m(\theta)=(W,A),\,\text{where,} \\
    \qquad [W]_{i,j} = 0 \text{ when }[E_W]_{i,j}=0,\,
    [W]_{i,j} = [\theta]_k,\,\text{ for some }k\le s_m, \text{ otherwise}\\
    \qquad [A]_{i,j} = 0 \text{ when }[E_A]_{i,j}=0,\,
    [A]_{i,j} = [\theta]_k,\,\text{ for some }k\le s_m, \text{ otherwise, uniquely}
\end{array}
\end{equation}
The details (lexicographical ordering, etc.) we leave out. Importantly, we have that $\mathbf{E}(\phi^m(\theta))$ presents a loss function for $\theta\in \mathbb{R}^{s_m}$.

With all the notation in place, we present the augmented CGVI with mixture weights by:
\begin{equation}\label{eq:cgvidbn}
    \min\limits_{P\in \mathcal{D},\mathbf{m}\in\Delta}
    \sum\limits_{m=1}^M \mathbf{m}_m\mathbb{E}_{P_m(\theta^m)}\left[ \mathbf{E}_N(\phi^m(\theta^m)\right]    
\end{equation}
where,
\begin{equation}\label{eq:weightdist}
P=\prod\limits_{m=1}^M P_m(\theta^m),\,\theta^m\in \mathcal{M}(\mathbb{R}^{s_m}) 
\end{equation}
that is, a distribution for $\{\theta^i\}_{i\in[M]}$ with each $P_m$ independent. Here $\mathcal{M}(\cdot)$ denotes a distribution over the argument space.

The measure constraint is defined as $\mathcal{D}=\prod^{\otimes}_{m\in[M]}\mathcal{D}_m$ with,
\begin{equation}\label{eq:uncertainset}
    \mathcal{D}_m = \left\{p(\theta)\in\mathcal{M}(\mathbb{R}^{s_m})\vert D_{\alpha}\left(p(\theta)\vert\vert \delta_{\Theta_m}\right)\le \epsilon\right\}
\end{equation}
Where $\delta_{\Theta_m}$ is a delta distribution on $[\Theta_m]_{\mathcal{I}_m}=\left\{[W]_{ij}:[E_W]_{ij}\neq 0\right\}\cup\left\{[A]_{ij}:[E_A]_{ij}\neq 0\right\}$. Thus $\mathcal{D}_m$ is meant to ensure that $\theta$, as sampled from distributions of posterior-maximizing weights for the potential function $\mathbf{E}_N$, is within a probability distance, the specific metric being defined by $D_{\alpha}$, of full measure at the original empirical solution $\Theta_m$, restricted to the dimensionality of $s_m$.


Due to the linearity of the expectation operator, we can see that we can perform the weight optimization and sampling offline for each structure, and then subsequently optimize the weight mixture. Formally:
\begin{equation}\label{eq:cgvihier}
    \begin{array}{rl}
\tilde{\Theta}^m \in \arg\min\limits_{P(\theta^m)\in\mathcal{D}_m} &\mathbb{E}_{P(\theta^m)}\left[\mathbf{E}_N(\phi^m(\theta^m))\right]
\\
\longrightarrow \mathbf{m} \sim & \mathbf{m}_i=\frac{\exp\left\{- \mathbb{E}_{\theta^i\sim P(\tilde{\Theta}^i)}\left[\left(\mathbf{E}_N(\phi^i(\theta^i))\right)\right]\right\} }{\sum\limits_{m=1}^M\exp\left\{- \mathbb{E}_{\theta^m\sim P(\tilde{\Theta}^m)}\left[\mathbf{E}_N(\phi^m(\theta^m))\right] \right\}}
    \end{array}
\end{equation}
Thus, we can perform the entire procedure, integer programming, sampling, and convex optimization, for every model $m\in[M]$, entirely independently and in parallel. Then, subsequently, we sample from the mixture as weighted by the marginal posterior for that mixture. Note that this would correspond to a noisy selection by the standard Bayesian Score Criterion as is standard for evaluating DBN structures~\cite{koller2009probabilistic}

\section{Algorithms}
\subsection{Optimization}
Now we discuss the specific procedures we use in order to solve the measure valued optimization problem,
\[
\tilde{\Theta}^m \in \arg\min\limits_{P(\theta^m)\in\mathcal{D}_m} \mathbb{E}_{P(\theta^m)}\left[\mathbf{E}_N(\phi^m(\theta^m))\right]
\]
for each model $m$. For this, we directly apply the procedures introduced for CGVI in~\cite{javeed2023risk}. To begin with, taking the specific form of~\eqref{eq:cgvioptgeneral} in our case yields the dual problem
\begin{equation}
    \min\limits_{\lambda\ge 0,\mu\in\mathbb{R}} \, \left\{\mu+\epsilon\lambda+\lambda\left(1+\frac{
    \mathbb{E}_{P(\tilde{\theta}^m)}\left[\mathbf{E}_N(\phi^m(\theta^m))\right]-\mu}{\beta}\right)^{\beta}-1\right\}
\end{equation}
Where $\beta<0$ is some parameter. This is an optimization problem convex in two variables $\lambda$ and $\mu$. However, it is not locally Lipschitz, thus care must be taken as far as optimization. Precise well-conditioned methods must be emphasized, which is possible with the low dimension.

The definition of the objective in the CGVI problem involves an expectation over$\Pi$, the prior distribution from which the divergence should be bounded, which in our case is $\delta_{\Theta_m}$, that is, the delta distribution centered at the IP parameter solution $\Theta_m$. And so the optimization problem to solve in our case is, for all $m\in[M]$
\begin{equation}\label{eq:cgvioptprob}
    \min\limits_{\lambda\ge 0,\mu\in\mathbb{R}} \, \left\{\mu+\epsilon\lambda+\lambda\left(1+\frac{
    \mathbf{E}^m_N-\mu}{\beta}\right)^{\beta}-1\right\}
\end{equation}
where $\mathbf{E}^m_N=\mathbf{E}(\Xi_m,\Theta_m;\mathcal{S})$


\subsection{Sampling}
The posterior, given optimal $\bar{\lambda}$ and $\bar{\mu}$ can be sampled from the distribution:
\begin{equation}\label{eq:posteriorcgvi}
    p(\theta^m) = \left(1+\frac{\mathbf{E}_N(\phi^m(\theta^m))-\bar{\mu}}{\beta\bar{\lambda}}\right)^{\beta-1}
\end{equation}
and alternatively,
\begin{equation}\label{eq:posteriorcgvigibbs}
\begin{array}{l}
    p(\theta) = \exp\left[\log\left\{\left(1+\frac{\mathbf{E}_N(\phi^m(\theta^m))-\bar{\mu}}{\beta\bar{\lambda}}\right)^{\beta-1}\right\}\right]\\ \quad 
    =\exp\left\{
    (\beta-1)\log\left[1+\frac{\mathbf{E}_N(\phi^m(\theta^m))-\bar{\mu}}{\beta\bar{\lambda}}\right]\right\}
\end{array}
\end{equation}

Recall again that
\[
\begin{array}{l}
\mathbf{E}\left(\phi^m(\theta^m)\right)=\mathbf{E}\left(E^{m}_W(\theta^m),E^{m}_A(\theta^m),W^m(\theta^m),A^m(\theta^m);\{X^{n,t}\}_{n=1,...,N,t=1,...,T}\right)
\\ =\left(\sum\limits_{n=1}^N\sum\limits_{t=1}^T\sum\limits_{i=1}^d\left([X_{n,t}]_i-\sum\limits_{j=1}^d [E^{m}_W(\theta^m)]_{j,i}W^m_{j,i}(\theta^m)[X_{n,t}]_j-\sum\limits_{l=1}^p\sum\limits_{j=1}^d [E^{m}_A(\theta^m)]_{l,j,i} A_{l,j,i}(\theta^m) [X_{m,t-l}]_j\right)^2\right)
\end{array}
\]
where the multiplication of the binary and continuous variables $E_W \cdot W$, etc. is redundant, and just noted for presentation.

\subsection{Complete Algorithm}
For completeness, now we describe the full procedure, incorporating all of the components of the algorithm described above. This is defined as Algorithm~\ref{alg:main}.

\begin{algorithm}
\begin{algorithmic}
\STATE \textbf{Input: } Data with $N$ trajectories of $T$ time steps each $\{X_{n,t}\}$. Number of total subsamples $M$, initial values $\mu_0,\lambda_0$, divergence parameter $\beta<0$, distance bound $\epsilon$
\FOR{$m=1,...,M$}
\STATE Select data sub-samples as by~\eqref{eq:datanotation} and Algorithm~\ref{alg:initip}, that is,
\[
\mathbf{S}(\{X_{n,t}\}_N,S)= \mathcal{S}^m=\{X_{i,t}\}_{i=n_m^1,...,n_m^S},\,\{n_m^1,...,n_m^S\}\sim \mathcal{U}\left\{\begin{array}{c} N \\ S \end{array}\right\}
\]
\STATE Solve~\eqref{eq:ip} with the data samples $\mathcal{S}^m=\{X_{i,t}\}_{i=n_m^1,...,n_m^S}$ to obtain the mixed-integer solution $\{\Theta^m,\Xi^m\}$
\STATE Let $\lambda=\lambda_0$ and $\mu=\mu_0$ and set \texttt{Converged} $\leftarrow$ \emph{FALSE}
\STATE Compute $\mathbf{E}^m_N=\mathbf{E}(\Xi_m,\Theta_m;\mathcal{S})$
\STATE Solve the optimization problem,
\[
\min\limits_{\mu\in\mathbb{R},\lambda\ge 0} \left\{\mu+\epsilon \lambda +\lambda\left(1+\frac{\mathbf{E}^m_N-\mu}{\beta}\right)^{\beta}-1\right\}
\]
to obtain $(\mu^*_m,\lambda^*_m)$

\STATE Sample~\eqref{eq:posteriorcgvi}, using the entire dataset $\{X_{i,t}\}_N$ and given $\lambda,\mu=(\mu^*_m,\lambda^*_m)$. After a burn in period, return a set of $Q$ samples $\{\Theta_1,\Theta_2,...,\Theta_Q\}$
\ENDFOR
\textbf{ and return} :
\begin{enumerate}
    \item Set of Structures $\{\Xi_m\}$ and weights $\{\Theta_m\}$
    \item Set of CGVI weight samples for each mixture $\{\Theta_q^m\}_{q\in[Q],m\in[M]}$
\item the mixture vector $\mathbf{m}^*$ as by~\eqref{eq:cgvihier}
\end{enumerate}
\STATE 
\textbf{Evaluate:}
\FOR{$r=1,...,R$}
\STATE Using a validation dataset $\mathcal{V}=\{X_{n,t}\}_{N',T}$, sample from:
\begin{equation}\label{eq:evalsamp}\begin{array}{l}
E^r = \mathbf{E}_V(\phi^m(\theta^m)):=\mathbf{E}(\Xi_m,\phi^m_W(\theta^m),\phi^m_A(\theta^m);\mathcal{V})\\  , \, (m,\theta^m)\sim  \mathbf{m}^*,\,\{\Theta_q^m\}_{q\in[Q]}
\end{array}
\end{equation}
with $\mathbf{E}_V$ denoting the loss on the validation dataset $\mathcal{V}$ and $\{\Theta_q^m\}_{q\in[Q]}$ is the sample chain at the optimal $(\mu_m^*,\lambda_m^*)$ solution
\ENDFOR
\RETURN : Evaluate the model based on the following values:
\begin{equation}\label{eq:results}
    \left\{\{E^r\}_{r\in[R]}\cup \{\mathbf{E}_V(\phi^m(\Theta_m))\}_{m\in[M]}\right\}
\end{equation}
where recall that $\Theta_m$ is the $m$'th integer programming solution for weights. 
\end{algorithmic}
\caption{Empirical Bayes for Dynamic Bayesian Network Learning Algorithm}
\label{alg:main}
\end{algorithm}
The evaluation sample~\eqref{eq:evalsamp} is a multinomial with coefficients given by the components of $\mathbf{m}$. It can be understood as the following simple operation: $\rho\sim\mathcal{U}[0,1]$, that is, a uniform random number  between 0 and 1, then $m=\arg\min \{i:\sum_{j=1}^i [\mathbf{m}]_j \le \rho \}$, and finally $\theta^m$ is one sample chosen uniformly at random from $\{\Theta_q^m\}_{q\in[Q]}$, the set of samples generated by model $m$ at the optimal $\bar{\lambda},\bar{\mu}$).

\section{Numerical Results}
\textcolor{red}{Note: this is a work on progress, and the numerical results reported here are exploratory}

Here we will report on the numerical results of the methods described above. Fundamentally, understanding and evaluating Bayesian models involves distinct intentions and best practices than optimized point estimates. In the Bayesian case, it is understood that the model will be uncertain, and the uncertainty being accurately sampled from indicates an appropriate algorithm. With many posteriors, most ergodic samplers achieve this. Thus the purpose of a Bayesian model is to adequately characterize the uncertain distribution regarding the parameters, and as long as the chain mixes, it is not meaningful to say a particular model is poor and well performing. Rather, it is meant to showcase the best you can do, as far as learning the reality from the data. 

One uses a Bayesian model similarly in inference, sampling from the mixture and weights and computing some quantity of interest, e.g.~$p(X_3(t+1)\vert X_2(t),X_5(t+1)) $ with other dependencies of the $X_3(t+1)$, for instance $X_4(t)$, are marginalized out. One can acquire a probability distribution of this inference from sampling the model. For causal inference, one can consider that if a certain conditional independence relationship exists among the most probable structures in the mixture, then this would correspond to a high probability causal conjecture, given the data.




We test Algorithm~\ref{alg:main} by performing a Train-Validation split. 

We do the following to illustrate the noise:
\begin{enumerate}
    \item Split the data into 70/30 test/validation split. 
    \item Perform Algorithm~\ref{alg:main}, and obtain an optimal mixture weight $m$ and set of samples of weights for each mixture $\{\Theta^{m_i}_q\}$
    \item Compute $ \mathbf{E}_N(m_i,\Theta^{m_i}_q)$ on the validation set for some set of samples
    \item Plot the histogram to estimate the density function of the loss given the data
    \item Indicate the loss the IP solutions achieved
\end{enumerate}

\begin{figure}
    \includegraphics[scale=1.2]{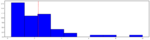}
    \includegraphics[scale=1.2]{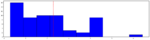}
    \caption{Representative histograms and empirical estimates for the validation loss}\label{fig:hist}
\end{figure}

In Figure~\ref{fig:hist} we see the histogram of the distribution of $\{E_r\}$, together with the values of several $E^m_V$ overlayed as vertical lines on the histogram density estimate. 

\section*{Acknowledgments}
The authors would like to thank and Ond{\v r}ej Ku{\v z}elka for his suggestions and discussion on this work. This work has received funding from the European Union’s Horizon Europe research and innovation programme under grant agreement No. 101084642.
\bibliographystyle{plain}
\bibliography{refs.bib}
\end{document}